\documentclass{article}
\usepackage{spconf,amsmath,graphicx}
\usepackage{array}
\usepackage{enumitem}
\usepackage{algorithm}
\usepackage{algpseudocode}
\setlist{nosep, leftmargin=14pt}
\usepackage{algpseudocode}
\usepackage{mwe} % to get dummy images
\usepackage[subtle]{savetrees}
\usepackage{url}
\title{An Attentive Representative Sample Selection Strategy Combined With Balanced Batch Training for Skin Lesion Segmentation}

 \name{Stephen Lloyd-Brown$^{1}$ Susan Francis$^{2}$ \textit{Caroline Hoad}$^{2}$ \textit{Penny Gowland}$^{2}$\ \textit{Karen Mullinger}$^{2}$ \textit{Andrew French}$^{1}$ \textit{Xin Chen}$^{1}$ }
 \address{$^{1}$ School of Computer Science, University of Nottingham, United Kingdom. \\
     $^{2}$ Sir Peter Mansfield Imaging Centre, University of Nottingham, United Kingdom}
\begin{document}
%\ninept
%

\maketitle
\begin{abstract}

\textit{© 2025 IEEE. Personal use of this material is permitted. Permission from IEEE must be obtained for all other uses, in any current or future media, including reprinting/republishing this material for advertising or promotional purposes, creating new collective works, for resale or redistribution to servers or lists, or reuse of any copyrighted component of this work in other works.} 
\par
An often overlooked problem in medical image segmentation research is the effective selection of training subsets to annotate from a complete set of unlabelled data. Many studies select their training sets at random, which may lead to suboptimal model performance, especially in the minimal supervision setting where each training image has a profound effect on performance outcomes. This work aims to address this issue. We use prototypical contrasting learning and clustering to extract representative and diverse samples for annotation. We improve upon prior works with a bespoke cluster-based image selection process. Additionally, we introduce the concept of unsupervised balanced batch dataloading to medical image segmentation, which aims to improve model learning with minimally annotated data. We evaluated our method on a public skin lesion dataset (ISIC 2018) and compared it to another state-of-the-art data sampling method. Our method achieved superior performance in a low annotation budget scenario. Github link: \url{https://github.com/ppysl3/nnunetv2_BalBatch}
\end{abstract}

\vspace*{-0.5\baselineskip}
\section{Introduction}

\label{sec:intro}

Medical image segmentation is an important tool used in both clinical and research settings. Fundamentally, it allows for regions of interest to be extracted from a complete medical image (e.g. highlighting a particular organ, tissue, or lesion). This improves the efficiency of downstream analysis. 
%Traditionally segmentation is performed by hand. Manual segmentation still represents the gold standard, 
In many clinical applications, segmentation is still performed manually, which is treated as a gold standard for decision-making and training automatic image segmentation models (e.g. deep learning based models \cite{hatamizadeh2021swin}). These deep learning based models can achieve superior performance if a large number of images with their annotated masks are available to train them.
However, the image annotation process is time consuming %\cite{mcgrath2020manual} 
and requires expert knowledge of the anatomy in question%\cite{rachmadi2018automatic}
, which increases expense. \par

\par
Large quantities and high quality of data are often the key to developing modern machine learning based methods. Compared to image segmentation methods themselves, training set sampling is a relatively understudied issue. The naive yet common approach is to select a subset at random from the training set. This approach is flawed 
as it increases the chance of selecting repetitive information, resulting in sub-optimal performance and wasted labor resources. As such, more effective sampling solutions are desirable. 
\par
One strategy that has become popular in recent years is using iterative active learning. Here the network queries samples to annotate throughout training to provide informative samples to the system. While this will ultimately end up reducing the annotation burden, its key flaw is the requirement to have an annotator on standby each time the network halts for additional annotation.  In the medical setting, it is not practical to query experts several times throughout many training cycles. As such, a method which requests all required samples for labelling in one go is desired.  \par
One-shot active learning (OSAL) aims to select a subset of data for annotation in one shot, without the need for repetitive querying. The use of OSAL to select representative training sets in medical image segmentation is severely understudied, with the author only identifying a handful of researchers working in this area. The seminal work of Zheng et al. \cite{zheng2019biomedical} made use of autoencoders, generative adversarial networks, and variational autoencoders as unsupervised feature extractors, followed by a clustering and sample selection stage. Their work shows that OSAL is a viable method for improving segmentation performance in the limited annotation budget setting. The core concepts were improved by contrastive annotation \cite{jin2022one}, which instead used more modern contrastive learning methods to generate more powerful representations. Their work demonstrates methods for selecting an optimal number of clusters, as well as how to sample from these clusters to form a training set, but uses MoCo \cite{he2020momentum} as a feature extractor, which has since been superseded by several other contrastive methods. Additionally, reliance solely on the silhouette coefficient to determine cluster quality can lead to unstable results. It also discards clustering information after sample selection has concluded.
\par
Performance could be improved across methods by incorporating clusters into the training process, ensuring that at each batch, samples from each cluster are viewed equally. This prevents reduced performance caused by random order loading, which can be detrimental at smaller test set sizes. This technique has been applied in the supervised setting \cite{huang2020balance} but to our knowledge has yet to be applied to OSAL. We hypothesise that it helps to achieve improved performance at no extra annotation or model training cost.
\par
In this paper, we aim to reduce the annotation burden by more attentively selecting samples to annotate compared to a random selection. We use prototypical contrastive learning (PCL) \cite{li2020prototypical} and K-means clustering \cite{bishop2006pattern} to construct representative clusters using a bespoke k-optimisation process. We then sample images from each cluster to label using a comprehensive sampling strategy. Balanced batch loading is adopted from supervised learning to our unsupervised setting, which aims to improve training performance compared to random batch loading. Our method was evaluated and compared to other methods using a public skin lesion dataset. The key contributions are summarized below.\\
\begin{itemize}

\item The adoption of PCL \cite{li2020prototypical} as an OSAL feature extractor, due to its improved performance in literature over other feature extractors.
\item An effective and consistent K-Optimisation method is proposed which more effectively determines the hyperparameter K for K-means clustering.
\item For the first time we adopt the idea of balanced batch loading from the supervised setting into OSAL for medical image segmentation to improve performance.

\end{itemize}
%\vspace*{-1.5\baselineskip}
\section{Methodology} 
\begin{figure}
    \centering
    \includegraphics[scale=0.0995]{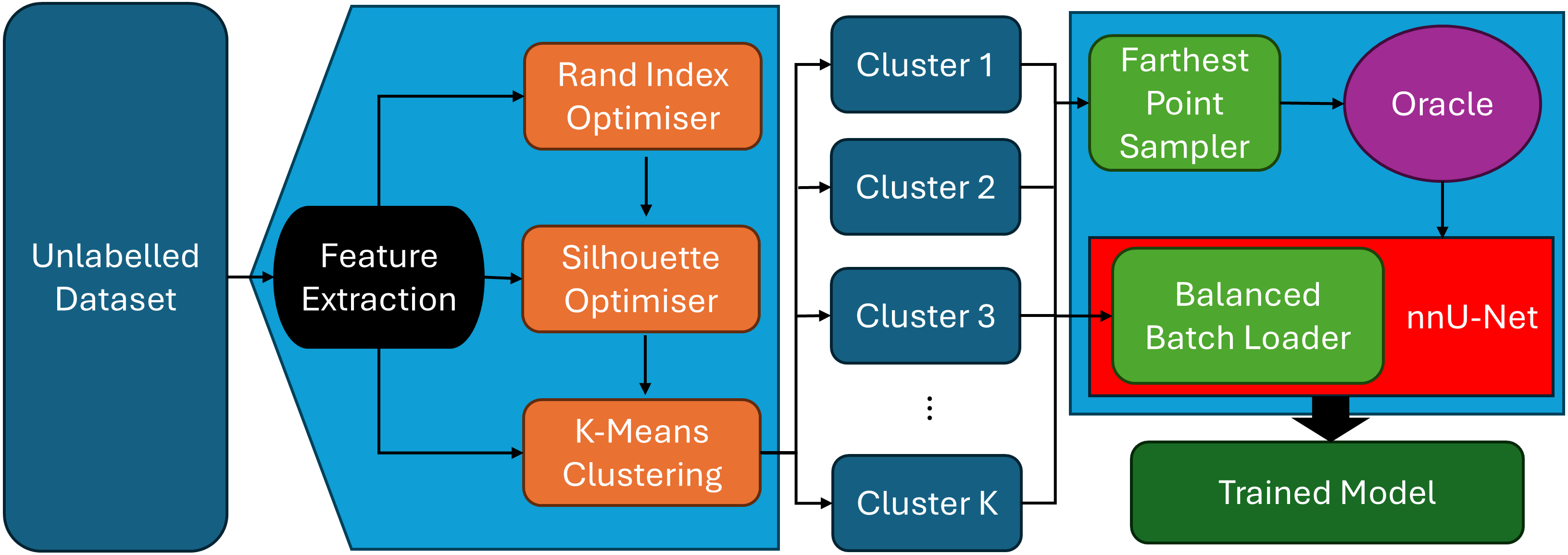}
    \caption{Diagram of our sample selection pipeline.}
    \label{fig:BalBatchDiagram}
\end{figure}

%\vspace*{-0.5\baselineskip}
\label{sec:Methodology}
An overview of our proposed pipeline is shown in Fig. \ref{fig:BalBatchDiagram}. Our method begins by taking a fully unlabeled dataset and passing it through
a feature extraction network based on the idea of contrastive learning. All data samples $N$ are clustered into $K$ different groups based on their extracted features using K-Means \cite{bishop2006pattern} clustering. $K$ is automatically determined by a bespoke optimization process. Subsequently, a fixed amount of data $D$ is sampled evenly from each cluster to form a training set 
$\{X\textsubscript{1}, X\textsubscript{2}, ... ,X\textsubscript{D}\}$
(where $D \ll N$), which is then proposed for obtaining their manual labels $\{Y\textsubscript{1}, Y\textsubscript{2}, ... ,Y\textsubscript{D}\}$. A balanced batch loader is also utilized to ensure balanced sampling from all clusters in each of the training epochs in the learning process. Finally, a deep convolutional neural network (DCNN) based image segmentation model is trained based on these selected data samples to achieve automatic segmentation of the unseen data. The details of each component are provided in the following subsections.
\vspace*{-0.5\baselineskip}

\subsection{Unsupervised Feature Extraction}
\vspace*{-0.5\baselineskip}

 Due to the assumed absence of labels, feature extraction must be performed in an unsupervised manner. 
 Contrastive networks typically work by passing images through two feature extractors, pulling features which originate from the same image together, while repelling features which originate from different images \cite{chen2020simple}\cite{he2020momentum}. There are many interpretations of this paradigm \cite{chen2020simple}\cite{wu2018unsupervised}\cite{he2020momentum}\cite{li2020prototypical}. One popular interpretation is Momentum Contrast (MoCo) \cite{he2020momentum}.
 Contrastive learning works best with a large number of samples to contrast against, and as such, MoCo follows prior authors \cite{wu2018unsupervised} in maintaining a queue of previously computed representations for this purpose. MoCo observes that older representations are less likely to be accurate, as the network has had time to update since encoding the oldest example.
 To solve this, MoCo introduces a slowly moving momentum encoder to encode the samples stored in the dictionary. This creates a powerful feature extractor. A key flaw with MoCo, and many other contrastive methods \cite{chen2020simple}, is that semantically similar images will be repelled in feature space, owing to the fact that their representations originate from different images. This can result in decreased performance, especially should you wish to cluster the images downstream. As such, prototypical contrastive learning (PCL) \cite{li2020prototypical} addresses this issue by adding a clustering stage to training.
 Here, representations are clustered into a number of clusters ({PCL\textunderscore K}), with representations within these clusters encouraged to be similar.
% which pulls representations from similar samples together. 
 The result is a state-of-the-art feature extractor, which is used on the training set. Extracted features are then passed to the next stage of the pipeline.

 \vspace*{-0.5\baselineskip}

\subsection{Balanced Sampling}
For the next stage of the pipeline extracted features are used in order to select a subset of data for annotation. K-means is an efficient and widely-used clustering method that can be applied to separating the data into representative groups. First, however, an appropriate number of clusters $K$ must be determined. 
\par Many methods, which aim to select an ideal number of clusters $K$, rely solely on silhouette coefficient \cite{rousseeuw1987silhouettes}\cite{jin2022one}. We observe however that selecting the most effective $K$ by silhouette coefficient is often not enough to ensure stability in image selection. Through repetition, clusters can be forced into unreproducible assignments which return high silhouette values, but are unreplicable between runs. We hypothesize that if a set of clusters truly exists within the data, we should be able to produce consistent cluster allocations for a given optimal $K$. Therefore we introduce an additional constraint using the rand index \cite{rand1971objective}. 
\par
The rand index was initially envisaged to evaluate consistency between two separate clustering methods. In our method, we re-purpose the rand index to compare multiple iterations of the same clustering method to itself. The rand index of a given clustering method's assignments, compared to ground truth, is often simplified in literature \cite{ismail2023shapedba} to the form in equation (\ref{eqn:rand}):
\begin{equation}
s=\frac{TP + TN}{TP + TN + FP + FN}
\label{eqn:rand}
\end{equation}
where TP and TN indicate agreement with the ground truth cluster assignments (true positive and true negative) and FP and FN indicate disagreement (false positive and false negative).
We use the slightly modified adjusted rand index \cite{hubert1985comparing} (ARI) which corrects for random chance.
We use this ARI to compare clustering assignments between iterations. 
In particular, for a given set of $K$, we calculate the average ARI of clustering into a given $K$ multiple times, where a high ARI indicates consistent cluster assignments for the given $K$.
The silhouette coefficient \cite{rousseeuw1987silhouettes} (equation (\ref{eqn:sil}), where $a$ is the mean intra-cluster distance, and $b$ is the mean inter-cluster distance.) is then used to further refine $K$.
\begin{equation}
s=\frac{b-a}{max(a,b)}
\label{eqn:sil}
\end{equation}
\noindent  
\par Finally, an issue arises in that the performance of K-means can be heavily influenced by the initial randomized
seed. As such, we improve upon prior works by calculating the silhouette coefficient for a given $K$ 100 times with different seeds. The final selection of $K$ is done first by evaluating $K$'s with high ARI's, and then further refining with the highest silhouette coefficient. This ensures that the resulting clusters have high repeatability and compactness. The process is described in algorithm \ref{alg:SelectKAlg}. All values of $K$ from 4 to 20 are considered, with the lower bound chosen so as not to trivialize the contribution of the rand index.
\begin{algorithm}
\caption{Multi-seed selection of $K$ based on rand index and silhouette coefficient}\label{alg:SelectKAlg}
\begin{algorithmic}
\ForAll{$K$}
    \For{ 100 random seeds}
        \State Run K-means with $Seed$ and $K$
        \State Calculate silhouette score
        \If{ New highest silhouette score for $K$}
            \State Store $Seed$ as $HiSeed$
            \State Store silhouette score as $HiScore$
        \EndIf 
        \If{ Not first run}
            \State Calculate ARI between the current 
            \State and the previous cluster assignments.
            \State Store ARI in an array.
        \EndIf 
    \EndFor
    \State{Calculate and store the average ARI and the corresponding }
    \State{$K$ value}
    \State{Store $HiSeed$ and $HiScore$ for that $K$}
\EndFor
%\For{ Highest 2 ARI's}
     \State{Rank the $K$ and $Seed$ pairs based on their corresponding average ARI values.}
    \State{Retrieve $K$ and $Seed$ with the highest silhouette coefficient from the top $T$ ranked candidates.}
%\EndFor
    \State \Return $K$ and $Seed$
\end{algorithmic}
\end{algorithm}
\par
\begin{figure}
    \centering
    \includegraphics[scale=0.2]{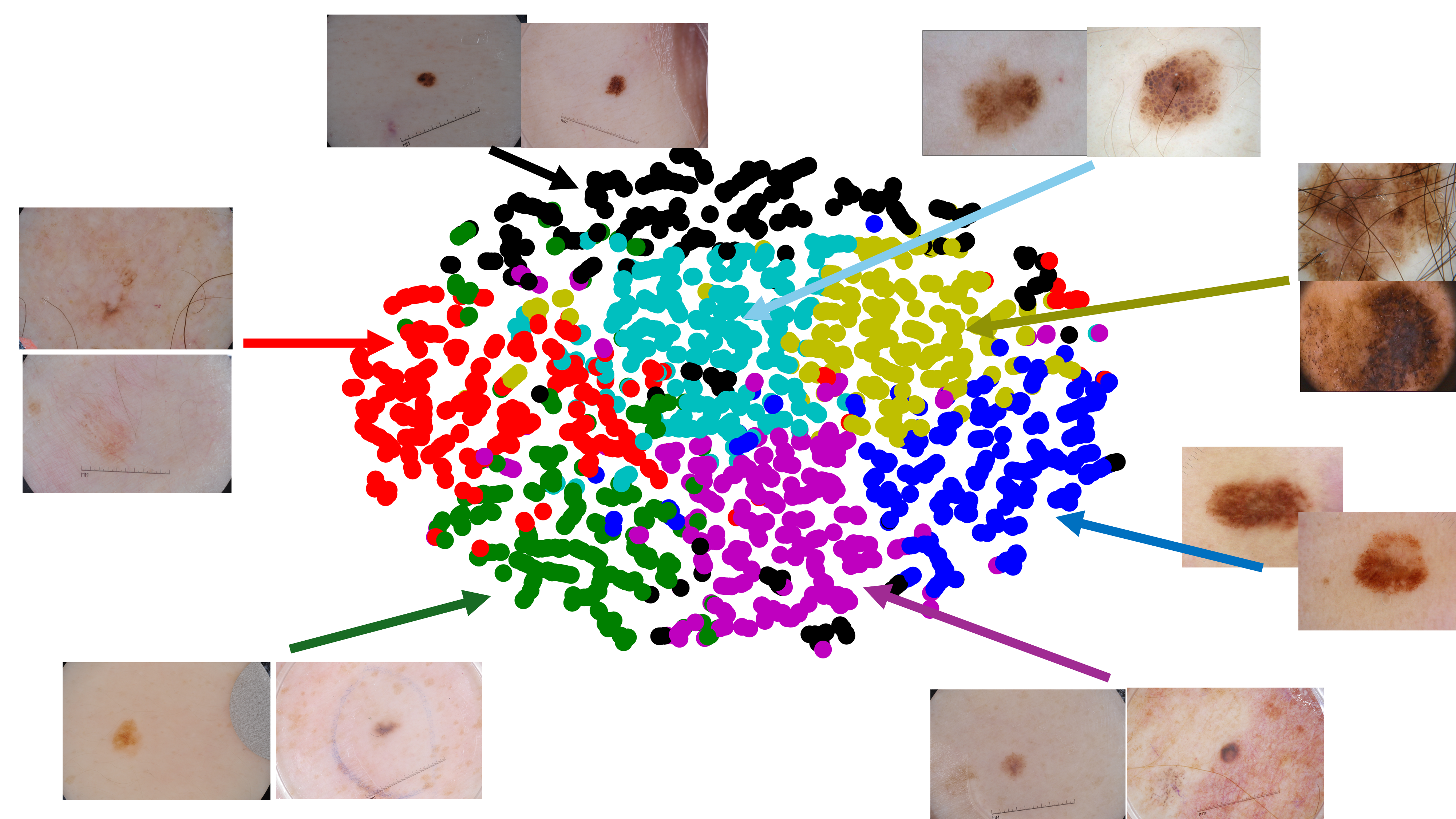}
    \caption{Diagram of ISIC 2018 feature space, projected into 2D space using TSNE \cite{van2008visualizing} color coded by cluster assignment, with clusters assigned using our pipeline discussed throughout section \ref{sec:Methodology}. We can see groupings based on size, opacity and color.}
    \label{fig:FeatureSpace}
\end{figure}

 Once an appropriate number of clusters $K$ has been selected, along with the respective seed, the entire unlabelled dataset is separated into clusters using K-means \cite{bishop2006pattern}. Figure \ref{fig:FeatureSpace} shows the learned clusters in the 2D feature space after applying TSNE \cite{van2008visualizing} dimensionality reduction method for the skin lesion dataset used in this paper. It can be seen that the proposed method is able to successfully group similar images into the same cluster. From each of these clusters, $N/K$ samples are extracted (where $N$ is the total affordable number of images for annotation) using farthest point sampling. In this scenario, the point closest to all other points is identified as the first to label. Subsequently, until the annotation budget is reached, the point farthest away from all currently selected points is iteratively selected for annotation. This results in a subset of images ready to be labeled and passed to the segmentation algorithm.
\vspace*{-0.5\baselineskip}
\subsection{Image Segmentation Model}
A commonly used DCNN-based medical image segmentation method is U-Net \cite{ronneberger2015u}, a U-shaped, fully convolutional, encoder-decoder style network, which demonstrates superior performance compared to prior methods \cite{ronneberger2015u}. It continues to be used as a base for many state-of-the-art segmentation systems \cite{hatamizadeh2021swin}. One such segmentation system is nnU-Net \cite{isensee2021nnu}. The defining feature of nnU-Net is the network's ability to optimize its own model architecture and hyperparameters. This allows nnU-Net to be trained fully automatically without explicitly requiring user parameter input. 

We chose nnU-Net as our segmentation network because the standardized nature of hyperparameter tuning removes training inconsistency as a factor in performance difference, allowing for greater confidence that differences in performance come as a direct result of changes to the pipeline.
\vspace*{-0.5\baselineskip}\textbf{}
\subsection{Balanced Batch Loading}
nnU-Net was built without the potential for cluster based data balancing in mind, and as such, provides no mechanism for loading examples in a specific order. Random batch sampling has the potential issue to suboptimally undersample from minority classes, resulting in poor performance during real-life model deployment. For instance, these minority classes could be life-critical cases that we can not afford to miss in disease screening. As such, we developed an unsupervised balanced batch loader for nnU-Net, which ensures that in each batch, an equal number of samples from each cluster is loaded. This ensures that at each batch, the system is able to see the full diversity of all clusters. 

\section{Method Evaluation}

\subsection{Datasets and Compared Methods}

Our proposed method was evaluated using the popular ISIC-2018 Skin Lesion Dataset \cite{gutman2016skin}. This dataset consists of 2594 training images, and 1000 testing images of skin lesions, along with their gold standard ground truth labels. The test images are kept in reserve and only used for the final evaluation.

In addition to our method detailed in section \ref{sec:Methodology}, we compared our method to three other sampling methods. We compared to our nearest competitor Contrastive Annotation \cite{jin2022one} (Competitor) and an average baseline of 4 randomly sampled sets (Random). We also performed an ablation study
using the modern SOTA feature extractor from MedSAM \cite{ma2024segment} (Ours - MedSAM) evaluating its viability as a standalone feature extractor, and demonstrating the importance of the cluster friendly feature extractor PCL.
An additional ablation study was performed without the use of the balanced batch dataloader (Ours - Ablation) in order to demonstrate the effectiveness of the balanced batch dataloader.
\subsection{Experimental Design} \label{ExpDes}
%Between all methods  strict annotation budget of 200 images (10\% of total data) was enforced.

For the feature selection stage our method employs PCL \cite{li2020prototypical}, which uses MoCo \cite{he2020momentum} as a backbone. Contrastive annotation \cite{jin2022one} uses MoCo as its feature extractor. In order to allow for a fair comparison, all shared parameters between these methods were kept the same. As such, we used a learning rate of 0.03, a batch size of 80, a dictionary of size 160, and a temperature of 0.2, keeping all other hyperparameters as used in the original paper. \par
For PCL specific hyperparameters, we set our number of PCL clusters ({PCL\textunderscore K}) to 200, as the literature suggests that overclustering is beneficial to training \cite{li2020prototypical}, and the network itself suggests having at least 10 times the amount of data as clusters. Both feature extraction networks are allowed to train for 200 epochs. The feature extractor from MedSAM \cite{ma2024segment} was initialised from frozen weights provided by the authors.\par
Our $K$ value optimisation algorithm ran for 100 random seeds as specified in algorithm \ref{alg:SelectKAlg}, with our pipeline consistently selecting $K$ to be 7. We set hyperparameter $T$ to 2.
Due to the enforcement of equal sampling of each cluster, the subset of images for labelling must be divisible by $K$. As such, with the MedSAM selecting $K=4$, and our method selecting $K=7$, we chose the common multiple 56 (2.16\% of total data) as our annotation interval.

For nnU-Net, barring the modification of our method using the balanced batch dataloader, all hyperparameters were left default for nnU-Net to automatically determine. These were done in a plan and preprocessing stage indifferent to downstream methodology. The sole exception to this was the batch size, which due to the need to sample evenly between clusters, must be a multiple of the number of clusters $K$. For simplicity, the batch size was manually set to $K$, which in our case was 7. nnU-Net was allowed to train for the default 1000 epochs. Each trained nnU-Net segmentation model was then evaluated on the full ISIC 2018 test set.

The initial feature extractor for each method was trained once only, with feature extraction providing consistent outputs owing to the frozen weights of the network at inference. This reduces the complexity of evaluating the later stages of the pipeline. 

\vspace*{-1\baselineskip}
\subsection{Results} \label{results}
\begin{figure}
    \centering
    \includegraphics[scale=0.6228]{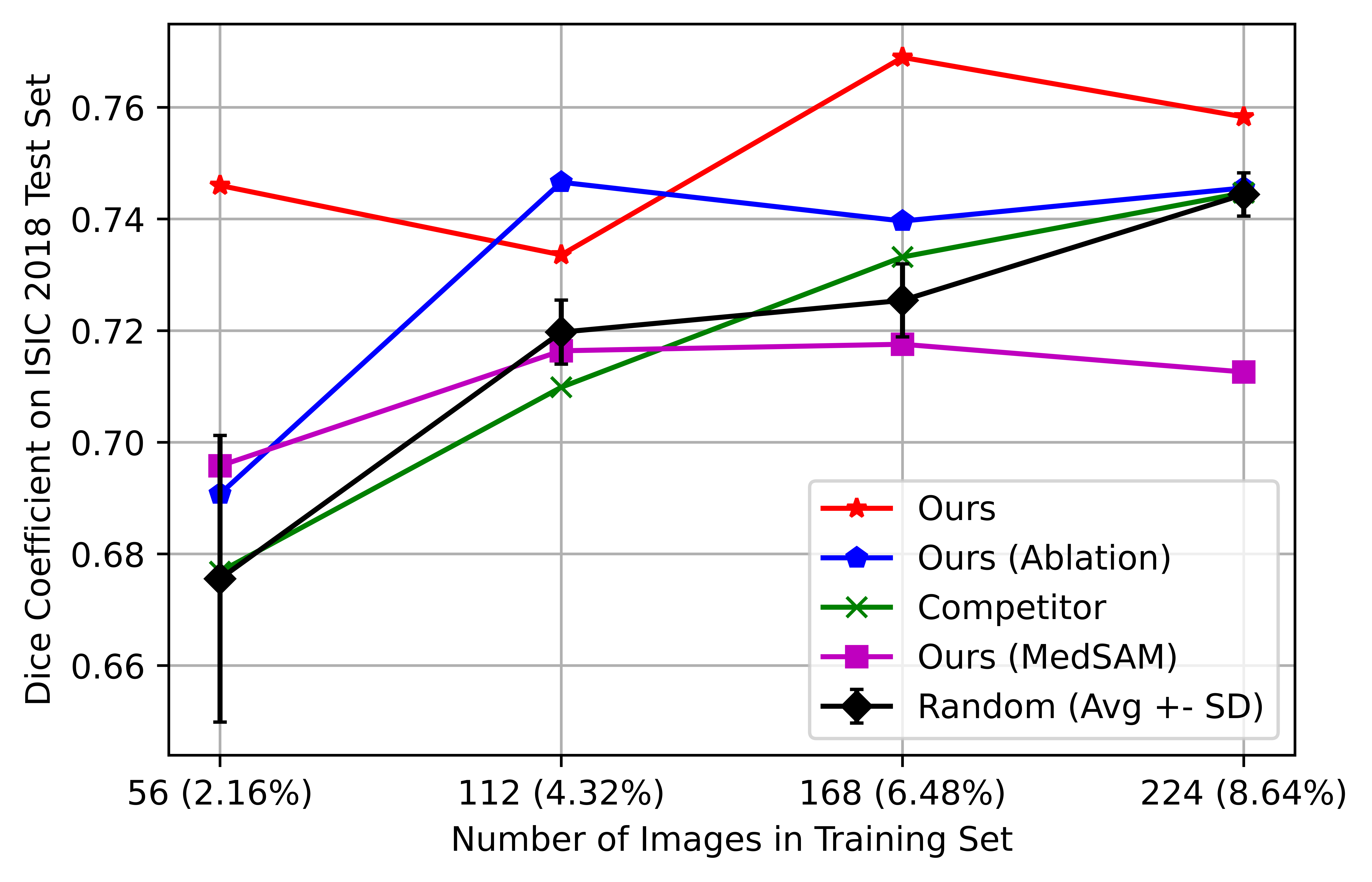}
    \caption{DICE Score reported per method and annotation budget. We compare our method, our method without balanced batch loading (Ablation), the SOTA method Contrastive Annotation \cite{jin2022one} (Competitor), our method using a non-clustering powerful feature extractor (MedSAM \cite{ma2024segment}) and random selection (4 runs).}
    \label{fig:Results}
\end{figure}
\vspace*{-0.5\baselineskip}
Our results reported in Fig. \ref{fig:Results} show a decisive improvement using our method (Ours) compared to the nearest SOTA (Competitor) \cite{jin2022one} and the popular Random Selection. Our ablation study (Ablation) shows the importance of the balanced batch loader, with it's inclusion often improving performance compared to its omission. Our study using MedSAM (MedSAM) \cite{ma2024segment} as a feature extractor shows a decrease in performance compared to our choice of PCL \cite{li2020prototypical}, showing the importance of a cluster oriented feature extractor.

\vspace*{-0.5\baselineskip}
\section{Discussion and Conclusions}
While this paper focuses primarily on selecting a subset of training data to annotate, it neglects the abundance of unannotated data that remains unselected. Future research would expand this work through the use of semi-supervised learning in order to leverage this unlabeled data to improve performance.

Further work on exploring different values of K, and reducing the computational load of clustering would further enhance this work.
Expansion to other datasets, particularly MRI, is a natural progression of this project. The core principles could also be adapted to non-imaging domains.
Finally, while this methodology omits common enhancements such as data augmentation in order to make testing more straightforward, an eventual clinical solution would fully leverage such methods to improve segmentation performance. 

\section{Compliance with ethical standards}
\label{sec:ethics}

This research study was conducted retrospectively using human subject data made available in open access. Ethical approval was not required as confirmed by the licence attached to the open access data.

\section{Acknowledgments}
\label{sec:acknowledgments}

Stephen Lloyd-Brown is funded by the Engineering and Physical Sciences Research Council. We are grateful for access to the University of Nottingham High Performance Computing service.

% ------------------------------------------------------------------------- 
\bibliographystyle{IEEEbib}
\bibliography{strings,refs}

\end{document}